\documentclass[11pt,a4paper]{article}
\pdfoutput=1  %
\usepackage[hyperref]{acl2020}
\usepackage{times}
\usepackage{latexsym}

\usepackage{microtype}

\usepackage{makecell}
\usepackage{tabularx}
\usepackage{enumitem}
\usepackage{amssymb}
\usepackage{amsmath}
\usepackage[normalem]{ulem}
\usepackage{graphicx}

\usepackage{url}

\usepackage{booktabs}

\aclfinalcopy %
\def\aclpaperid{1826} %

\title{Probing Linguistic Features of Sentence-Level Representations in Neural Relation Extraction}

\author{Christoph Alt ~~~~~ Aleksandra Gabryszak ~~~~~
Leonhard Hennig \\
\mbox{}\\
German Research Center for Artificial Intelligence (DFKI)\\
Speech and Language Technology Lab \\
\texttt{\{christoph.alt, aleksandra.gabryszak, leonhard.hennig\}@dfki.de}}

\date{}

\begin{document}
\maketitle

\newcommand{\inserttacredresulttable}{
    \begin{table}[ht!]\centering
  \small
        \begin{tabular}{lccc}
        \toprule
                                   &  P   &  R   &  F1  \\
        \midrule
         BoE                       & 50.0 & 32.6 & 39.4 \\
        \midrule
         CNN                       & 72.3 & 45.5 & 55.9 \\
         \quad + ELMo              & \underline{\textbf{73.8}} & 48.9 & 58.8 \\
         \quad + BERT $\downarrow$ & 71.9 & 51.1 & 59.7 \\
         \quad + BERT $\uparrow$   & 69.8 & 54.3 & 61.0 \\
         CNN $\mathrm{\otimes}$               & 67.2 & 53.4 & 59.5 \\
         \quad + ELMo              & 72.3 & 53.8 & 61.7 \\
         \quad + BERT $\downarrow$ & 69.0 & \textbf{62.0} & 65.3 \\
         \quad + BERT $\uparrow$   & 71.9 & 61.1 & \textbf{66.1} \\
         \midrule
         Bi-LSTM                   & 53.3 & 57.4 & 55.3 \\
         \quad + ELMo              & 65.1 & 58.8 & 61.8 \\
         \quad + BERT $\downarrow$ & 65.3 & 59.9 & 62.5 \\
         \quad + BERT $\uparrow$   & 65.2 & 61.2 & 63.1 \\
         Bi-LSTM $\mathrm{\otimes}$           & 62.5 & 63.4 & 62.9 \\
         \quad + ELMo              & 63.3 & 64.9 & 64.1 \\
         \quad + BERT $\downarrow$ & 64.9 & \textbf{66.0} & 65.4 \\
         \quad + BERT $\uparrow$   & \textbf{68.3} & 64.0 & \textbf{66.1} \\
         \midrule
         GCN                       & 65.4 & 51.1 & 57.4 \\
         \quad + ELMo              & 66.2 & 58.5 & 62.1 \\
         \quad + BERT $\downarrow$ & 66.1 & 59.9 & 62.9 \\
         \quad + BERT $\uparrow$   & 66.2 & 57.4 & 61.5 \\
         GCN $\mathrm{\otimes}$               & 68.1 & 59.8 & 63.7 \\
         \quad + ELMo              & \textbf{68.5} & 62.6 & 65.4 \\
         \quad + BERT $\downarrow$ & 68.1 & 64.5 & \textbf{66.3} \\
         \quad + BERT $\uparrow$   & 66.6 & \textbf{65.3} & 65.9 \\
         \midrule
         S-Att.                    & 56.9 & 58.3 & 57.6 \\
         \quad + ELMo              & 64.4 & 65.0 & 64.7 \\
         \quad + BERT $\downarrow$ & 60.6 & 67.6 & 63.9 \\
         \quad + BERT $\uparrow$   & 63.5 & 64.1 & 63.8 \\
         S-Att. $\mathrm{\otimes}$            & 65.0 & 66.8 & 65.9 \\
         \quad + ELMo              & 64.0 & 69.4 & 66.6 \\
         \quad + BERT $\downarrow$ & 64.0 & \underline{\textbf{69.7}} & 66.7 \\
         \quad + BERT $\uparrow$   & \textbf{69.2} & 64.7 & \underline{\textbf{66.9}} \\
        \bottomrule
        \end{tabular}
    \caption{Relation extraction test set performance on TACRED. $\uparrow$ and $\downarrow$ indicate the cased and uncased version of BERT, $\mathrm{\otimes}$ models with entity masking.}
    \label{tab:test_performance_tacred}

    \end{table}
}

\newcommand{\inserttacredprobingtable}{
    \begin{table*}[ht!]\centering
    \footnotesize
    \begingroup
    \setlength{\tabcolsep}{4pt}
        \begin{tabular}{@{}lcccccccccccccc|c@{}}
        \toprule
                               &  \thead{Type\\Head}  &  \thead{Type\\Tail}  &  \thead{Sent\\Len}  &  \thead{Arg\\Dist}  &  \thead{Arg\\Ord}  &  \thead{Ent\\Exist}  &  \thead{PosL\\Head}  &  \thead{PosR\\Head}  &  \thead{PosL\\Tail}  &  \thead{PosR\\Tail}  &  \thead{Tree\\Dep}  &  \thead{SDP\\Dep}  &  \thead{GR\\Head}  &  \thead{GR\\Tail} & \thead{F1\\score} \\
        \midrule
            Majority vote             &   66.4 &   33.5 &    14.5 &   14.8 &  54.7 & 51.0 &    22.8 &    23.0 &    26.9 &    20.0 &     23.7 &    28.4 &    58.4 &    75.2 &    - \\
            Length                    &   66.4 &   33.5 &   \textbf{100.0} &   13.8 &  54.8 & 59.4 &    18.6 &    24.7 &    26.9 &    20.1 &     \textbf{30.5} &    29.6 &    58.4 &    75.2 &    - \\
            ArgDist                   &   66.4 &   33.5 &    16.5 &  \textbf{100.0} &  54.7 & 77.5 &    14.9 &    23.0 &    26.9 &    19.8 &     23.8 &    35.3 &    58.4 &    75.2 &    - \\
            BoE                       &   77.7 &   47.6 &    61.1 &   22.6 &  97.3 & 66.5 &    33.7 &    41.5 &    32.5 &    36.3 &     29.8 &    31.0 &    66.3 &    77.4 & 39.4 \\
            \midrule
            CNN                       &   94.0 &   85.8 &    47.6 &   88.1 &  98.8 & \textbf{84.5} &    70.7 &    76.1 &    84.0 &    86.5 &     28.5 &    44.0 &    78.0 &    88.6 & 55.9 \\
            \quad + ELMo              &   \textbf{97.0} &   \textbf{90.2} &    48.7 &   91.7 &  99.1 & 84.3 &    \textbf{76.1} &    \textbf{81.2} &    \textbf{86.6} &    \textbf{90.1} &     28.3 &    45.0 &    \textbf{82.8} &    \textbf{91.9} & 58.8 \\
            \quad + BERT $\downarrow$ &   95.9 &   88.8 &    44.7 &   46.0 &  93.8 & 79.9 &    64.7 &    74.4 &    80.8 &    88.4 &     29.4 &    41.0 &    77.7 &    90.0 & 59.7 \\
            \quad + BERT $\uparrow$   &   96.1 &   88.8 &    48.0 &   43.7 &  91.9 & 80.0 &    56.9 &    70.3 &    80.1 &    87.5 &     28.0 &    41.3 &    75.0 &    89.6 & 61.0 \\
            CNN $\mathrm{\otimes}$    &   84.2 &   60.9 &    46.4 &   58.3 &  94.3 & 81.5 &    44.3 &    50.9 &    54.4 &    63.9 &     27.7 &    40.0 &    68.5 &    82.0 & 59.5 \\
            \quad + ELMo              &   82.8 &   69.8 &    47.4 &   75.6 &  98.1 & 82.9 &    54.2 &    60.2 &    65.4 &    77.3 &     28.7 &    42.4 &    71.9 &    85.0 & 61.7 \\
            \quad + BERT $\downarrow$ &   87.6 &   80.3 &    50.9 &   29.3 &  83.2 & 72.4 &    39.3 &    46.1 &    67.7 &    80.7 &     30.1 &    36.9 &    67.1 &    87.4 & 65.3 \\
            \quad + BERT $\uparrow$   &   87.2 &   79.3 &    50.6 &   25.3 &  78.3 & 69.8 &    39.6 &    42.9 &    59.9 &    77.5 &     30.3 &    35.1 &    65.6 &    86.9 & 66.1 \\
            \midrule
            Bi-LSTM                   &   93.4 &   81.2 &    42.0 &   47.9 &  \textbf{99.4} & 79.2 &    41.2 &    50.8 &    50.6 &    68.4 &     28.7 &    41.7 &    69.3 &    85.2 & 55.3 \\
            \quad + ELMo              &   96.4 &   89.6 &    27.9 &   47.0 &  97.9 & 80.9 &    47.8 &    52.5 &    67.2 &    72.6 &     25.2 &    42.8 &    72.1 &    90.0 & 61.8 \\
            \quad + BERT $\downarrow$ &   96.0 &   87.3 &    31.0 &   45.5 &  99.1 & 78.8 &    46.1 &    55.6 &    61.7 &    71.3 &     26.6 &    42.7 &    72.2 &    87.7 & 62.5 \\
            \quad + BERT $\uparrow$   &   96.0 &   87.7 &    28.6 &   45.3 &  97.7 & 80.4 &    48.0 &    50.9 &    61.4 &    67.4 &     25.1 &    42.3 &    70.8 &    87.0 & 63.1 \\
            Bi-LSTM $\mathrm{\otimes}$&   81.9 &   71.4 &    27.6 &   35.6 &  90.6 & 73.2 &    36.1 &    40.5 &    59.3 &    66.4 &     25.7 &    38.4 &    64.6 &    85.3 & 62.9 \\
            \quad + ELMo              &   82.8 &   50.7 &    30.6 &   19.7 &  73.4 & 65.0 &    32.0 &    35.9 &    37.9 &    41.8 &     28.0 &    32.2 &    63.0 &    79.5 & 64.1 \\
            \quad + BERT $\downarrow$ &   82.3 &   77.9 &    34.1 &   25.6 &  87.6 & 68.4 &    32.5 &    36.7 &    61.5 &    64.7 &     27.6 &    35.1 &    66.6 &    86.0 & 65.4 \\
            \quad + BERT $\uparrow$   &   81.7 &   79.6 &    30.2 &   21.3 &  81.1 & 67.0 &    30.6 &    33.8 &    55.9 &    55.1 &     27.3 &    34.2 &    64.1 &    84.9 & 66.1 \\
            \midrule
            GCN                       &   93.0 &   81.9 &    18.8 &   35.5 &  86.0 & 74.4 &    48.6 &    48.8 &    51.2 &    52.3 &     24.0 &    \textbf{49.9} &    74.2 &    85.9 & 57.4 \\
            \quad + ELMo              &   96.3 &   86.2 &    18.7 &   29.3 &  77.5 & 74.0 &    50.4 &    52.0 &    48.9 &    51.7 &     23.2 &    47.4 &    77.1 &    86.9 & 62.1 \\
            \quad + BERT $\downarrow$ &   96.0 &   85.2 &    20.7 &   31.2 &  83.6 & 74.2 &    48.6 &    52.4 &    47.4 &    50.4 &     23.9 &    48.7 &    74.4 &    85.3 & 62.9 \\
            \quad + BERT $\uparrow$   &   96.3 &   85.7 &    21.4 &   32.9 &  84.3 & 75.3 &    50.1 &    54.6 &    48.6 &    52.5 &     24.5 &    49.2 &    76.3 &    85.8 & 61.5 \\
            GCN $\mathrm{\otimes}$    &   87.6 &   67.4 &    18.1 &   33.1 &  81.6 & 72.8 &    36.8 &    51.1 &    44.8 &    48.8 &     24.1 &    47.3 &    73.2 &    83.0 & 63.7 \\
            \quad + ELMo              &   92.7 &   68.6 &    18.6 &   26.4 &  76.8 & 71.4 &    41.9 &    50.4 &    43.6 &    45.1 &     23.8 &    47.1 &    76.3 &    83.9 & 65.4 \\
            \quad + BERT $\downarrow$ &   93.5 &   71.5 &    22.0 &   33.3 &  88.5 & 73.8 &    44.9 &    50.6 &    44.7 &    47.7 &     24.4 &    49.1 &    72.6 &    82.3 & 66.3 \\
            \quad + BERT $\uparrow$   &   93.4 &   72.0 &    23.7 &   33.2 &  90.4 & 73.9 &    42.8 &    50.1 &    44.0 &    48.3 &     24.9 &    48.0 &    72.9 &    83.0 & 65.9 \\
            \midrule
            S-Att.                    &   89.9 &   81.8 &    22.7 &   32.8 &  75.7 & 78.1 &    34.1 &    38.9 &    40.8 &    44.8 &     26.1 &    38.2 &    60.7 &    81.1 & 57.6 \\
            \quad + ELMo              &   96.6 &   87.8 &    24.9 &   30.6 &  74.1 & 79.1 &    36.0 &    41.4 &    39.2 &    44.1 &     26.4 &    37.9 &    64.1 &    83.4 & 64.7 \\
            \quad + BERT $\downarrow$ &   96.2 &   87.0 &    25.9 &   31.4 &  75.6 & 76.5 &    35.3 &    40.8 &    39.8 &    44.4 &     25.4 &    39.1 &    61.8 &    81.3 & 63.9 \\
            \quad + BERT $\uparrow$   &   96.5 &   87.3 &    26.1 &   32.6 &  76.8 & 78.0 &    34.7 &    40.9 &    40.0 &    44.0 &     25.7 &    38.1 &    62.2 &    81.7 & 63.8 \\
            S-Att. $\mathrm{\otimes}$ &   79.5 &   56.5 &    29.0 &   44.3 &  91.2 & 79.5 &    29.6 &    43.0 &    36.1 &    60.3 &     26.1 &    39.6 &    64.7 &    79.5 & 65.9 \\
            \quad + ELMo              &   78.2 &   44.4 &    25.1 &   31.5 &  72.3 & 77.1 &    31.6 &    37.5 &    34.4 &    34.8 &     26.2 &    36.7 &    62.1 &    75.9 & 66.6 \\
            \quad + BERT $\downarrow$ &   82.4 &   66.9 &    36.2 &   33.2 &  74.9 & 76.8 &    32.0 &    37.6 &    38.0 &    41.3 &     27.4 &    37.6 &    63.0 &    79.8 & 66.7 \\
            \quad + BERT $\uparrow$   &   80.0 &   69.0 &    31.9 &   32.8 &  78.6 & 76.6 &    30.3 &    34.2 &    37.5 &    39.2 &     27.0 &    38.2 &    60.4 &    79.9 & \textbf{66.9} \\
        \bottomrule
        \end{tabular}
    \caption{TACRED probing task accuracies and model F1 scores on the test set. $\uparrow$ and $\downarrow$ indicate the cased and uncased version of BERT, $\mathrm{\otimes}$ models with entity masking. Probing task classification is performed by a logistic regression on the representations $s_j$ of all sentences in the dataset.}
    \label{tab:probing_task_results_tacred}
     \endgroup
    \end{table*}
}

\newcommand{\insertsemevalresulttable}{
    \begin{table}[ht!]\centering
    \small
        \begin{tabular}{lccc}
        \toprule
                                   &  P   &  R   &  F1  \\
        \midrule
        BoE                       & 53.7 & 60.8 & 55.7 \\
        \midrule
        CNN                       & 81.8 & 78.9 & 80.2 \\
        \quad + ELMo              & 87.5 & 81.6 & 84.4 \\
        \quad + BERT $\downarrow$ & \textbf{89.5} & \underline{\textbf{83.4}} & \underline{\textbf{86.3}} \\
        \quad + BERT $\uparrow$   & 88.9 & 83.3 & 86.0 \\
        \midrule
        Bi-LSTM                      & 82.7 & 77.9 & 80.1 \\
        \quad + ELMo              & 87.3 & 80.6 & 83.7 \\
        \quad + BERT $\downarrow$ & \textbf{88.3} & \textbf{83.2} & \textbf{85.6} \\
        \quad + BERT $\uparrow$   & 87.5 & 83.0 & 85.1 \\
        \midrule
        GCN                       & 81.9 & 77.5 & 79.6 \\
        \quad + ELMo              & 86.1 & \textbf{82.6} & 84.2 \\
        \quad + BERT $\downarrow$ & \textbf{89.2} & \textbf{82.6} & \textbf{85.7} \\
        \quad + BERT $\uparrow$   & 87.6 & 81.4 & 84.3 \\
        \midrule
        S-Att.                    & 83.3 & 77.7 & 80.2 \\
        \quad + ELMo              & 87.7 & 79.9 & 83.6 \\
        \quad + BERT $\downarrow$ & \underline{\textbf{89.7}} & \textbf{81.9} & \textbf{85.6} \\
        \quad + BERT $\uparrow$   & 88.9 & 81.5 & 84.9 \\
        \bottomrule
        \end{tabular}
    \caption{Relation extraction test set performance on SemEval. $\uparrow$ and $\downarrow$ indicate the cased and uncased version of BERT. Due to the small dataset size, we report the mean across 5 randomly initialized runs.}
    \label{tab:test_performance_semeval}
    \end{table}
}

\newcommand{\insertsemevalprobingtable}{
    \begin{table*}[ht]\centering
    \footnotesize
        \begin{tabular}{@{}lcccccccccccc|c@{}}
        \toprule
                                   &  \thead{Type\\Head}  &  \thead{Type\\Tail}  &  \thead{Sent\\Len}  &  \thead{Arg\\Dist}  &  \thead{PosL\\Head}  &  \thead{PosR\\Head}  &  \thead{PosL\\Tail}  &  \thead{PosR\\Tail}  &  \thead{Tree\\Dep}  &  \thead{SDP\\Dep}  &  \thead{GR\\Head}  &  \thead{GR\\Tail} & \thead{F1\\score} \\
        \midrule
            Majority vote             &   22.0 &   21.3 &    25.7 &   42.1 &    62.1 &    39.3 &    38.3 &    34.0 &     25.4 &    67.2 &    37.3 &    80.9 &    - \\
            Length                    &   25.8 &   24.7 &   \textbf{100.0} &   42.1 &    62.1 &    39.1 &    38.3 &    46.3 &     \textbf{44.3} &    67.2 &    40.6 &    80.9 &    - \\
            ArgDist                   &   23.6 &   22.3 &    25.7 &  \textbf{100.0} &    62.1 &    43.7 &    37.9 &    35.3 &     26.2 &    67.8 &    45.4 &    80.9 &    - \\
            BoE                       &   58.5 &   58.0 &    82.4 &   84.8 &    65.1 &    66.1 &    49.2 &    72.5 &     44.1 &    69.8 &    65.4 &    83.6 & 55.7 \\
            \midrule
            CNN                       &   76.1 &   76.2 &    34.9 &   87.5 &    66.0 &    85.8 &    74.2 &    73.1 &     34.1 &    72.1 &    70.3 &    89.1 & 80.2 \\
            \quad + ELMo              &   81.3 &   81.8 &    38.1 &   88.5 &    70.0 &    89.0 &    79.5 &    76.4 &     35.5 &    71.8 &    75.1 &    90.9 & 84.4 \\
            \quad + BERT $\downarrow$ &   \textbf{83.9} &   \textbf{84.1} &    55.9 &   90.2 &    74.0 &    89.3 &    81.2 &    \textbf{84.6} &     41.3 &    73.1 &    76.8 &    90.6 & \textbf{86.3} \\
            \quad + BERT $\uparrow$   &   83.4 &   83.7 &    54.3 &   90.4 &    \textbf{74.4} &    \textbf{89.4} &    \textbf{82.0} &    82.8 &     42.0 &    73.0 &    78.3 &    90.8 & 86.0 \\
            \midrule
            Bi-LSTM                   &   77.1 &   77.0 &    50.5 &   74.9 &    63.8 &    75.9 &    61.8 &    68.5 &     41.3 &    70.3 &    69.2 &    87.7 & 80.1 \\
            \quad + ELMo              &   81.5 &   81.8 &    41.1 &   66.6 &    62.8 &    71.8 &    59.3 &    64.5 &     37.5 &    70.1 &    70.0 &    87.6 & 83.7 \\
            \quad + BERT $\downarrow$ &   83.6 &   83.7 &    41.8 &   61.5 &    62.7 &    68.9 &    57.9 &    63.0 &     37.1 &    70.8 &    67.4 &    86.7 & 85.6 \\
            \quad + BERT $\uparrow$   &   82.5 &   82.8 &    41.8 &   66.0 &    63.1 &    70.8 &    58.6 &    64.3 &     37.7 &    71.0 &    68.9 &    87.5 & 85.1 \\
            \midrule
            GCN                       &   75.4 &   75.5 &    35.0 &   81.5 &    68.5 &    87.5 &    71.2 &    55.5 &     35.5 &    \textbf{80.3} &    76.3 &    91.7 & 79.6 \\
            \quad + ELMo              &   80.7 &   80.8 &    32.2 &   68.1 &    68.3 &    83.4 &    65.8 &    53.2 &     34.4 &    75.8 &    80.0 &    91.1 & 84.2 \\
            \quad + BERT $\downarrow$ &   82.5 &   83.0 &    42.5 &   66.5 &    73.6 &    84.7 &    69.2 &    66.3 &     38.9 &    77.2 &    82.1 &    91.0 & 85.7 \\
            \quad + BERT $\uparrow$   &   81.5 &   81.9 &    42.7 &   67.3 &    73.8 &    85.1 &    69.6 &    67.8 &     39.6 &    77.6 &    \textbf{84.2} &    \textbf{91.9} & 84.3 \\
            \midrule
            S-Att.                    &   77.4 &   77.6 &    34.2 &   50.0 &    62.1 &    56.2 &    49.8 &    47.1 &     35.9 &    67.9 &    54.2 &    84.1 & 80.2 \\
            \quad + ELMo              &   80.7 &   81.3 &    33.1 &   46.2 &    62.0 &    53.9 &    49.1 &    45.7 &     34.7 &    68.1 &    54.9 &    84.4 & 83.6 \\
            \quad + BERT $\downarrow$ &   83.4 &   83.3 &    31.0 &   45.3 &    62.1 &    51.8 &    48.4 &    44.7 &     33.0 &    67.8 &    53.3 &    83.6 & 85.6 \\
            \quad + BERT $\uparrow$   &   82.8 &   82.8 &    30.6 &   46.1 &    62.1 &    52.7 &    48.2 &    44.4 &     33.6 &    67.9 &    54.6 &    84.1 & 84.9 \\
        \bottomrule
        \end{tabular}
    \caption{SemEval probing task accuracies and model F1 scores on the test set. $\uparrow$ and $\downarrow$ indicate the cased and uncased version of BERT.  Probing task classification is performed by a logistic regression on the representations $s_j$ of all sentences in the dataset.}
    \label{tab:probing_task_results_semeval}
    \end{table*}
}

\newcommand{\inserttacredarchitectureresulttable}{
\begin{table*}[ht]\centering
\footnotesize
\begin{tabular}{@{}lcccccccccccccc@{}}
\toprule
                       &  \thead{Type\\Head}  &  \thead{Type\\Tail}  &  \thead{Sent\\Len}  &  \thead{Arg\\Dist}  &  \thead{Arg\\Ord}  &  \thead{Ent\\Exist}  &  \thead{PosL\\Head}  &  \thead{PosR\\Head}  &  \thead{PosL\\Tail}  &  \thead{PosR\\Tail}  &  \thead{Tree\\Dep}  &  \thead{SDP\\Dep}  &  \thead{GR\\Head}  &  \thead{GR\\Tail}  \\
        \midrule
 Majority vote                      &  66.4   &  33.5   &   14.5   &  14.8   &  54.7  & 51.0  &   22.8   &   23.0   &   26.9   &   20.0   &   23.7    &   28.4   &   58.4   &   75.2   \\
 Length                             &  66.4   &  33.5   &  100.0   &  13.8   &  54.8  & 59.4  &   18.6   &   24.7   &   26.9   &   20.1   &   30.5    &   29.6   &   58.4   &   75.2   \\
 ArgDist                            &  66.4   &  33.5   &   16.5   &  100.0  &  54.7  & 77.5  &   14.9   &   23.0   &   26.9   &   19.8   &   23.8    &   35.3   &   58.4   &   75.2   \\
 BoE                                &  77.7   &  47.6   &   61.1   &  22.6   &  97.3  & 66.5  &   33.7   &   41.5   &   32.5   &   36.3   &   29.8    &   31.0   &   66.3   &   77.4   \\
        \midrule
        \textbf{Encoder} \\
        \midrule
Bi-LSTM       &   88.9 &   76.2 &    31.2 &   35.4 &  89.7 & 73.8 &    39.6 &    44.6 &    55.6 &    61.8 &     26.7 &    38.5 &    68.1 &    85.2 \\
CNN          &   89.4 &   78.9 &    47.8 &   60.3 &  93.2 & 80.1 &    55.3 &    62.4 &    70.8 &    81.4 &     28.9 &    41.0 &    72.9 &    87.2 \\
GCN          &   93.9 &   76.9 &    19.8 &   31.4 &  82.7 & 73.6 &    45.2 &    51.7 &    46.5 &    49.3 &     23.9 &    48.3 &    74.7 &    84.3 \\
Self-attention         &   87.4 &   72.6 &    27.7 &   33.6 &  77.4 & 77.7 &    32.9 &    39.3 &    38.2 &    44.1 &     26.3 &    38.2 &    62.4 &    80.3 \\
        \midrule
        \textbf{Entity Masking} \\
        \midrule
No          &   95.4 &   86.4 &    30.2 &   44.0 &  90.3 & 78.6 &    48.2 &    53.5 &    58.2 &    64.0 &     26.2 &    42.6 &    71.5 &    86.4 \\
Yes           &   84.7 &   65.7 &    32.2 &   33.7 &  82.2 & 72.5 &    37.4 &    43.3 &    48.2 &    53.9 &     26.8 &    39.0 &    67.0 &    82.4 \\
        \midrule
        \textbf{Word Repr.} \\
        \midrule
Word vectors       &   87.4 &   69.3 &    30.2 &   42.0 &  87.2 & 75.6 &    40.8 &    46.7 &    49.1 &    56.1 &     26.3 &    41.0 &    67.8 &    82.6 \\
ELMo         &   89.5 &   70.6 &    30.7 &   40.5 &  83.5 & 75.2 &    43.8 &    49.3 &    51.6 &    56.4 &     26.4 &    40.3 &    70.2 &    83.9 \\
BERT          &   90.6 &   81.4 &    32.5 &   34.0 &  86.5 & 74.8 &    41.6 &    47.4 &    55.5 &    61.1 &     26.7 &    40.6 &    68.8 &    85.3 \\
\bottomrule
\end{tabular}
    \caption{TACRED average probing task accuracies on models grouped by encoder architecture, by training with and without entity masking, and by input word representations. As we built our probing tasks on the SentEval toolkit, classification is performed by a logistic regression, taking the RE task-specific pre-learned sentence representations as input.}
    \label{tab:probing_task_results_tacred}
\end{table*}
}

\newcommand{\insertsemevalarchitectureresulttable}{
    \begin{table*}[ht]\centering
    \footnotesize
        \begin{tabular}{@{}lcccccccccccc@{}}
        \toprule
                                   &  \thead{Type\\Head}  &  \thead{Type\\Tail}  &  \thead{Sent\\Len}  &  \thead{Arg\\Dist}  &  \thead{PosL\\Head}  &  \thead{PosR\\Head}  &  \thead{PosL\\Tail}  &  \thead{PosR\\Tail}  &  \thead{Tree\\Dep}  &  \thead{SDP\\Dep}  &  \thead{GR\\Head}  &  \thead{GR\\Tail}  \\
        \midrule
 Majority vote             &  22.0   &  21.3   &   25.7   &  42.1   &   62.1   &   39.3   &   38.3   &   34.0   &   25.4    &   67.2   &   37.3   &   80.9   \\
 Length                    &  25.8   &  24.7   &  100.0   &  42.1   &   62.1   &   39.1   &   38.3   &   46.3   &   44.3    &   67.2   &   40.6   &   80.9   \\
 ArgDist                   &  23.6   &  22.3   &   25.7   &  100.0  &   62.1   &   43.7   &   37.9   &   35.3   &   26.2    &   67.8   &   45.4   &   80.9   \\
 BoE                       &  58.5   &  58.0   &   82.4   &  84.8   &   65.1   &   66.1   &   49.2   &   72.5   &   44.1    &   69.8   &   65.4   &   83.6   \\
        \midrule
        \textbf{Encoder} \\
        \midrule
Bi-LSTM       &   81.2 &   81.3 &    43.8 &   67.3 &    63.1 &    71.9 &    59.4 &    65.1 &     38.4 &    70.5 &    68.9 &    87.4 \\
CNN          &   81.2 &   81.4 &    45.8 &   89.1 &    71.1 &    88.3 &    79.2 &    79.2 &     38.2 &    72.5 &    75.1 &    90.3 \\
GCN          &   80.0 &   80.3 &    38.1 &   70.8 &    71.1 &    85.2 &    69.0 &    60.7 &     37.1 &    77.8 &    80.7 &    91.4 \\
Self-attention         &   81.1 &   81.3 &    32.2 &   46.9 &    62.1 &    53.7 &    48.9 &    45.5 &     34.3 &    67.9 &    54.2 &    84.1 \\
        \midrule
        \textbf{Entity Masking} \\
        \midrule
No          &   80.7 &   81.3 &    33.1 &   46.2 &    62.0 &    53.9 &    49.1 &    45.7 &     34.7 &    68.1 &    54.9 &    84.4 \\
Yes           &   80.9 &   81.1 &    40.4 &   70.0 &    67.2 &    76.1 &    65.1 &    63.7 &     37.2 &    72.5 &    70.7 &    88.5 \\
        \midrule
        \textbf{Word Representation} \\
        \midrule
Word vectors       &   76.5 &   76.6 &    38.6 &   73.5 &    65.1 &    76.3 &    64.2 &    61.0 &     36.7 &    72.7 &    67.5 &    88.2 \\
ELMo         &   81.0 &   81.4 &    36.1 &   67.3 &    65.8 &    74.5 &    63.4 &    60.0 &     35.5 &    71.4 &    70.0 &    88.5 \\
BERT          &   83.0 &   83.2 &    42.6 &   66.7 &    68.2 &    74.1 &    64.4 &    64.7 &     37.9 &    72.3 &    70.7 &    88.2 \\
\bottomrule
\end{tabular}
\caption{SemEval 2010 Task 8 average probing task accuracies on models grouped by encoder architecture, by training with and without entity masking, and by input word representations. Probing task classification is performed by a logistic regression, taking the RE task-specific pre-learned sentence representations as input.}
\label{tab:probing_task_results_semeval}
\end{table*}
}

\begin{abstract}

Despite the recent progress, little is known about the features captured by state-of-the-art neural relation extraction (RE) models. Common methods encode the source sentence, conditioned on the entity mentions, before classifying the relation. However, the complexity of the task makes it difficult to understand how encoder architecture and supporting linguistic knowledge affect the features learned by the encoder. We introduce 14 probing tasks targeting linguistic properties relevant to RE, and we use them to study representations learned by more than 40 different encoder architecture and linguistic feature combinations trained on two datasets, TACRED and SemEval 2010 Task 8. We find that the bias induced by the architecture and the inclusion of linguistic features are clearly expressed in the probing task performance. For example, adding contextualized word representations greatly increases performance on probing tasks with a focus on named entity and part-of-speech information, and yields better results in RE. In contrast, entity masking improves RE, but considerably lowers performance on entity type related probing tasks.

\end{abstract}

\section{Introduction}
\label{Introduction}
\begin{figure}[t!]
    \centering
    \includegraphics[width=\linewidth]{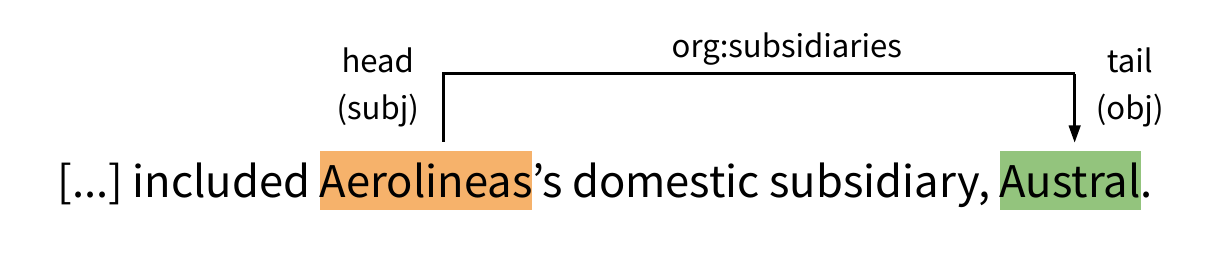}
    \caption{Example relation from TACRED. The sentence contains the relation \emph{org:subsidiaries} between the head and tail entities `Aerolineas' and `Austral'.}
    \label{fig:re_example}
\end{figure}

Relation extraction (RE) is concerned with extracting relationships between entities mentioned in text, where relations correspond to semantic categories such as \emph{org:founded\_by}, \emph{person:spouse}, or \textit{org:subsidiaries} (Figure \ref{fig:re_example}). Neural models have shown impressive results on this task, achieving state-of-the-art performance on standard datasets like SemEval2010 Task 8~\cite{dos_santos_classifying_2015,wang_relation_2016,lee_semantic_2019}, TACRED~\cite{zhang_graphco_2018, alt_improving_2019, peters-etal-2019-knowledge, joshi-etal-2019-spanbert}, and NYT~\cite{lin_neural_2016,vashishth_reside_2018, alt-etal-2019-fine}. The majority of models implement an encoder architecture to learn a fixed size representation of the input, e.g.\ a sentence, which is passed to a classification layer to predict the target relation label.

These good results suggest that the learned representations capture linguistic and semantic properties of the input that are relevant to the downstream RE task, an intuition that was previously discussed for a variety of other NLP tasks by \citet{conneau_probing_2018}. However, it is often unknown which exact properties the various models have learned. Our aim is to pinpoint the information a given RE model is relying on, in order to improve model performance as well as to diagnose errors.

A general approach to model introspection is the use of \emph{probing tasks}. Probing tasks \cite{shi_does_2016, adi_fine-grained_2017}, or diagnostic classifiers, are a well established method to analyze the presence of specific information in a model's latent representations, e.g.\ in machine-translation \cite{belinkov-etal-2017-neural}, language modeling \cite{giulianelli-etal-2018-hood}, and sentence encoding \cite{conneau_probing_2018}. For each probing task, a classifier is trained on a set of representations, and its performance measures how well the information is encoded. The probing task itself is typically selected in accordance with the downstream task, e.g.\ an encoder trained on RE may be probed for the entity type of a relation argument. If the classifier correctly predicts the type, it implies the encoder retains entity type information in the representations, which also directly inform the relation prediction. The simplicity of this approach makes it easier to pinpoint the information a model is relying on, as opposed to probing the downstream task directly.

Our goal in this paper is to understand which features of the input a model conditioned on relation extraction has learned as useful for the task, in order to be able to better interpret and explain model predictions.
Relation extraction literature is rich with information about useful features for the task~\cite{zhou_exploring_2005,mintz_distant_2009,surdeanu_customizing_2011}. Consequently, our initial question is whether and how good the sentence representations learned by state-of-the-art neural RE models encode these well-known features, such as e.g.\ argument entity types, dependency path or argument distance features. Another question is how the prior imposed by different encoding architectures, e.g.\ CNN, RNN, Graph Convolutional Network and Self-Attention, affects the features stored in the learned sentence representations. Finally, we would like to understand the effect of additional input features on the learned sentence representations. These include explicit semantic and syntactic knowledge like entity information and grammatical role, and as recently proposed, contextualized word representations such as ELMo~\cite{peters_deepcw_2018} and BERT~\cite{devlin_bert_2018}. 
We therefore significantly extend earlier work on probing tasks as follows:

\begin{itemize}[topsep=0pt,noitemsep]
    \item Following the framework of~\citet{conneau_probing_2018}, we propose a set of 14 probing tasks specifically focused on linguistic properties relevant to relation extraction.
    \item We evaluate four encoder architectures, also in combination with supporting linguistic knowledge, on two datasets, TACRED~\cite{zhang_position_aware_2017} and SemEval 2010 Task 8~\cite{hendrickx_semeval2010t8_2010}, for a total of more than 40 variants.
    \item We follow up on this analysis with an evaluation on the proposed probing tasks to establish a connection between task performance and captured linguistic properties.
    \item To facilitate further research and wider adoption, we open-source our relation extraction framework\footnote{\url{https://github.com/DFKI-NLP/RelEx}} based on AllenNLP~\cite{gardner_allennlp_2018}, and REval\footnote{\url{https://github.com/DFKI-NLP/REval}}, a framework extending the SentEval toolkit~\cite{conneau_senteval_2018} with our probing tasks.
\end{itemize}

\section{Probing Tasks}
This section introduces the probing tasks we use to evaluate the learned sentence representations. We base our work on the setup and tasks introduced by \citet{conneau_probing_2018}, but focus on probing tasks related to relation extraction. We therefore adopt some of the tasks they propose, and introduce new probing tasks specifically designed for RE. As in their work, the probing task classification problem requires only single sentence embeddings as input (as opposed to, e.g., sentence and word embeddings, or multiple sentence representations). This fits the standard RE setup quite well, where the task is typically to classify the relation(s) expressed between a pair of entity mentions in a single sentence. While we focus on supervised relation extraction, this setup is also applicable in a distantly supervised RE setting, where state-of-the-art approaches are often based on passing sentence representations to a bag-level classifier that computes classification label(s) over all sentences for a given entity pair~\cite{mintz_distant_2009,lin_neural_2016}. Similar to \citet{conneau_probing_2018}, we also aim to address a set of linguistic properties related to relation extraction ranging from simple surface phenomena (e.g.\ relation argument distance) to syntactic information (e.g.\ parse tree depth and argument ordering) and semantic information (e.g.\ the entity types of relation arguments). We use the standard training, validation, test split of the original TACRED dataset for RE and probing task experiments. For SemEval we reuse test and use 10\% of the training set for validation. For TACRED we use the provided named entity, part-of-speech, and dependency parsing information, and parse SemEval with the Stanford Parser (2018-10-05 version)~\cite{manning_stanford_2014}.
\paragraph{Surface information}
These tasks test whether sentence embeddings capture simple surface properties of sentences they encode. The sentence length (\textbf{SentLen}) task, introduced by~\citet{adi_fine-grained_2017}, predicts the number of tokens in a sentence. We group sentences into n = 10 bins (TACRED, 7 bins for SemEval) by length, selecting bin widths so that training sentences are distributed approximately uniformly across bins, and treat SentLen as a n-way classification task. Our next probing task, argument distance (\textbf{ArgDist}), predicts the number of tokens between the two relation arguments. Similar to SentLen, we group sentences into 10 bins (5 for SemEval) by relative distance. Inspired by a common feature in classical RE~\cite{surdeanu_customizing_2011}, we also test if any named entity exists between the two relation arguments (\textbf{EntExist}), treating it as a binary classification problem. Addressing this task requires the encoder to produce a sentence embedding that (at least partially) represents the inner context of the relation arguments.

\paragraph{Syntactic information}
Syntactic information is highly relevant for relation extraction. Many RE approaches utilize e.g.\ dependency path information~\cite{bunescu_shortest_2005,krause_large-scale_2012,mintz_distant_2009}, or part-of-speech tags~\cite{zhou_exploring_2005,surdeanu_customizing_2011}. We therefore include the tree depth  task (\textbf{TreeDepth}) described by \citet{conneau_probing_2018}. This task tests whether an encoder can group sentences by the depth of the longest path from root to any leaf. We group tree depth values into 10 (TACRED, SemEval 7) approximately uniformly distributed classes, ranging from from depth 1 to depth 15. To account for shortest dependency path (SDP) information, we include an SDP tree depth task (\textbf{SDPTreeDepth}), which tests if the learned sentence embedding stores information about the syntactical link between the relation arguments. Again, we group SDP tree depth values into bins, in this case only 6 (4) classes, since the SDP trees are generally more shallow than the original sentence dependency parse tree.
The argument ordering task (\textbf{ArgOrd}) tests if the head argument of a relation occurs before the tail argument in the token sequence. An encoder that successfully addresses this challenge captures  some information about syntactic structures where the order of a relation's arguments is inverted, e.g.\ in  constructions such as ``The acquisition of Monsanto by Bayer'', as compared to default constructions like ``Bayer acquired Monsanto''. 
We also include 4 tasks that test for the part-of-speech tag of the token directly to the left or right of the relation's arguments: \textbf{PosHeadL}, \textbf{PosHeadR}, \textbf{PosTailL}, \textbf{PosTailR}. These tasks test whether the encoder is sensitive to the immediate context of an argument. Some relation types, e.g.\ \emph{per:nationality} or \emph{org:top\_member}, can often be identified based on the immediate argument context, e.g.\ ``US \emph{president-NN} Donald Trump'', or ``Google \emph{'s-POSS CEO-NN} Larry Page''. Representing this type of information in the sentence embedding should be useful for the relation classification. 

\paragraph{Argument information}
Finally, we include probing tasks that require some understanding of what each argument denotes. The argument entity type tasks (\textbf{TypeHead, TypeTail}) ask for the entity tag of the head, and respectively the tail, argument. Entity type information is highly relevant for relation extraction systems since it strongly constrains the set of possible relation labels for a given argument pair. We treat these tasks as multi-class classification problems over the set of possible argument entity tags (see Section~\ref{subsec:datasets}).

Our last task concerns the grammatical function of relation arguments. The grammatical role tasks (\textbf{GRHead, GRTail}) ask for the role of each argument, as given by the dependency label connecting the argument and its syntactic head token. The motivation is that the subject and object of verbal constructions often correspond to relation arguments for some relation types, e.g.\ ``Bayer acquired Monsanto''. We currently test for four roles, namely \emph{nsubj, nsubjpass, dobj} and \emph{iobj}, and group all other dependency labels into the \emph{other} class. Note that there are other grammatical relations that may be of interest for relation extraction, for example possessive modifiers (``Google's Larry Page''), compounds (``Google CEO Larry Page''), and appositions (``Larry Page, CEO of Google'').

\section{Experiment Setup}
\label{sec:re_models}
This section first introduces the four sentence encoding architectures we consider for evaluation~(\S\ref{subsec:sent_encoders}), followed by a description of the supporting linguistic knowledge we evaluate: entity masking and contextualized word representations (\S\ref{subsec:add_input_features}). We also introduce the two datasets we use for training the relation extraction models and probing the sentence representations (\S\ref{subsec:datasets}).
\begin{figure*}[t!]
    \centering
    \includegraphics[width=\linewidth]{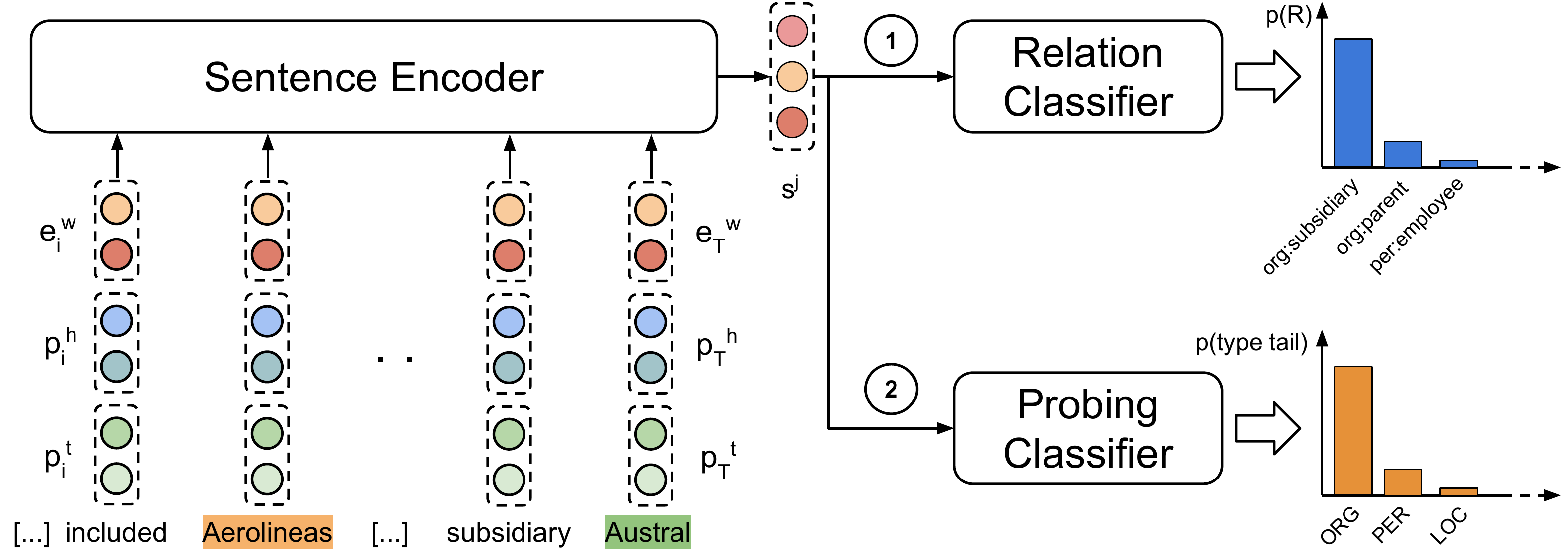}
    \caption{Probing task setup. In the first step, we train a RE model (sentence encoder and relation classifier) on a dataset $D$. In the second step, we fix the encoder and for each probing task train a classifier on the encoder representations $\{s_j\}_{j=1,\ldots,|D|}$ of all sentences in $D$. The probing classifier performance indicates how well the sentence representations encode the information probed by the classifier, e.g.\ the entity type of the tail relation argument.}
    \label{fig:probing_arch}
\end{figure*}

\subsection{Sentence Encoders}
\label{subsec:sent_encoders}
Generally, methods in relation extraction follow the sequence to vector approach, encoding the input (often a single sentence) into a fixed-size representation, before applying a fully connected relation classification layer (Figure \ref{fig:probing_arch}). A single input is represented as a sequence of T tokens $\{w_t\}_{t=1, \ldots, T}$, and the spans\ $(head_{start},\ head_{end})$ and\ $(tail_{start},\ tail_{end})$ of the two entity mentions in question. We focus our evaluation on four widely used approaches that have shown to perform well on RE. For all architectures we signal the position of \textit{head} and \textit{tail} by the relative offset to each token $w_i$ as a positional embedding $p^h_{i} \in \mathbb{R}^c$ and $p^t_{i} \in \mathbb{R}^c$ concatenated to the input token representation $e^t_i = [e^w_i,\ p^h_i,\ p^t_i]$, where ${e^w_i} \in \mathbb{R}^d$ is the token embedding.

\paragraph{CNN}
 We follow the work of \citet{zeng_relationcv_2014} and \citet{nguyen_recnn_2015}, who both use a convolutional neural network for relation extraction. Their models encode the input token sequence $\{w_t\}_{t=1, \ldots, T}$ by applying a series of 1-dimensional convolutions of different filter sizes, yielding a set of output feature maps $M_f$, followed by a max-pooling operation that selects the maximum values along the temporal dimension of $M_f$ to form a fixed-size representation. 

\paragraph{Bi-LSTM max} Similar to ~\citet{zhang_relationcv_2015} and ~\citet{zhang_position_aware_2017}, we use a Bi-LSTM to encode the input sequence. A Bi-LSTM yields a sequence of hidden states $\{h_t\}_{t=1, \ldots, T}$, where $h_t$ is a concatenation $[h^f_t, h^b_t]$ of the states of a forward LSTM $h^f$ and a backward LSTM $h^b$. Similar to the CNN, we use max pooling across the temporal dimension to obtain a fixed-size representation\footnote{We considered taking the final hidden state but found max pooling to perform superior.}.

\paragraph{GCN} Graph convolutional networks~\cite{kipf_graphconv_2016} adapt convolutional neural networks to graphs. Following the approach of~\citet{zhang_graphco_2018}, we treat the input token sequence $\{w_t\}_{t=1, \ldots, T}$ as a graph consisting of T nodes, with an edge between $w_i$ and $w_j$, if there exists a dependency edge between the two tokens. We convert the dependency tree into a $T \times T$ adjacency matrix, after pruning the graph to the shortest dependency path between \textit{head} and \textit{tail}. A L-layer GCN applied to $\{w_t\}_{t=1, \ldots, T}$ yields a sequence of hidden states $\{h_t\}_{t=1, \ldots, T}$ contextualized on neighboring tokens with a graph distance of at most L. Forming a fixed size representation is done by max pooling over the temporal dimension and local max pooling over the tokens $\{w_t\}$, for $t \in [head_{start}, \ldots, head_{end}]$ and similar for $t \in [tail_{start}, \ldots, tail_{end}]$.

\paragraph{Multi-Headed Self-Attention} Similar to the Transformer~\cite{vaswani_attention_2017}, we compute a sequence of contextualized representations $\{h_t\}_{t=1, \ldots, T}$ by applying L layers of multi-headed self-attention to the input token sequence $\{w_t\}_{t=1, \ldots, T}$. The representation ${h_t}$ of ${w_t}$ is computed as a weighted sum of a projection $V$ of the input tokens, with respect to the scaled, normalized dot product of $Q$ and $K$, which are also both linear projections of the input with the procedure repeated for each attention head. A fixed-size representation is obtained by taking the final state ${h_T}$ at the last layer L. %

\subsection{Supporting Linguistic Knowledge}
\label{subsec:add_input_features}
Adding additional lexical, syntactic, and semantic input features to neural RE approaches has been shown to considerably improve performance~\cite{zeng_relationcv_2014, zhang_position_aware_2017, zhang_graphco_2018}. Features include e.g.\ casing, named entity, part-of-speech and dependency information. Most recently, pre-learned contextualized word representations (deep language representations) emerged, capturing syntactic and semantic information useful to a wide range of downstream tasks~\cite{peters_deepcw_2018, radford_improvinglu_2018, devlin_bert_2018}. We therefore evaluate the effect of adding explicit named entity and grammatical role information (through entity masking) on our pre-learned sentence representations, and compare it to adding contextualized word representations computed by ELMo~\cite{peters_deepcw_2018} and BERT~\cite{devlin_bert_2018} as additional input features.

\paragraph{Entity Masking}
Entity masking has been shown to provide a significant gain for RE performance on the TACRED dataset~\cite{zhang_position_aware_2017} by replacing each entity mention with a combination of its entity type and grammatical role (subject or object). It limits the information about entity mentions available to a model, possibly preventing overfitting to specific mentions and forcing the model to focus more on the context.

\paragraph{ELMo} Embeddings from Language Models, as introduced by~\citet{peters_deepcw_2018}, are an approach to compute contextualized word representations by applying a pre-learned, two-layer Bi-LSTM neural network to an input token sequence $\{w_t\}_{t=1, \ldots, T}$. ELMo operates on a character level and is pre-trained with the forward and backward direction as a separate unidirectional language model. It yields a representation $h_i = [h^f_i,\ h^b_i]$ for each token $w_i$, with $h^f_i$ conditioned on the preceding context $\{w_t\}_{t=1, \ldots, i-1}$ and independently $h^b_i$, conditioned on the succeeding context $\{w_t\}_{t=i+1, \ldots, T}$.

\paragraph{BERT} Bidirectional Encoder Representations from Transformers~\cite{devlin_bert_2018} improves upon methods such as ELMo and the OpenAI Generative Pre-trained Transformer (GPT)~\cite{radford_improvinglu_2018} by using a masked language model that allows for jointly training forward and backward directions. Compared to ELMo, BERT operates on word-piece input and is based on the self-attentive Transformer architecture~\cite{vaswani_attention_2017}. It computes a representation for a token $w_i$ jointly conditioned on the preceding $\{w_t\}_{t=1, \ldots, i-1}$ and succeeding context $\{w_t\}_{t=i+1, \ldots, T}$.

\subsection{Datasets}
\label{subsec:datasets}
Table \ref{tab:dataset_stats} shows key statistics of the TACRED and SemEval datasets. TACRED is approximately 10x the size of SemEval 2010 Task 8, but contains a much higher fraction of negative training examples, making classification more challenging.
\begin{table}[ht!]
    \centering
    \footnotesize
    \begin{tabular}{@{}l c c c@{}}
        \toprule
        Dataset & \# Relations & \# Examples &  Neg. examples \\
        \midrule
        SemEval & 19 & ~~10,717 & 17.4\% \\
        TACRED & 42 & 106,264 & 79.5\% \\
        \bottomrule
    \end{tabular}
    \caption{Comparison of datasets used for evaluation}
    \label{tab:dataset_stats}
\end{table}

\paragraph{TACRED} The \textit{TAC} \textit{R}elation \textit{E}xtraction \textit{D}ataset\footnote{\url{https://catalog.ldc.upenn.edu/LDC2018T24}}~\cite{zhang_position_aware_2017} contains 106k sentences with entity mention pairs collected from the TAC KBP\footnote{\url{https://tac.nist.gov/2017/KBP/index.html}} evaluations. 
Sentences are annotated with person- and organization-oriented relation types, e.g.\ \emph{per:title}, \emph{org:founded} and \emph{no\_relation} for negative examples. In contrast to the SemEval dataset the entity mentions are typed with subjects classified into person and organization and objects categorized into 16 fine-grained classes (e.g., date, location, title). As per convention, we report our results as micro-averaged F1 scores.

\paragraph{SemEval 2010 Task 8} The SemEval 2010 Task~8 dataset\footnote{\url{http://www.kozareva.com/downloads.html}}~\cite{hendrickx_semeval2010t8_2010} is a standard benchmark for binary relation classification, and contains 8,000 sentences for training and 2,717 for testing. Sentences are annotated with a pair of untyped nominals and one of 9 directed semantic relation types, such as\ \emph{Cause-Effect}, \emph{Entity-Origin} as well as the undirected \emph{Other} type to indicate no relation, resulting in 19 distinct types in total. We follow the official convention and report macro-averaged F1 scores with directionality taken into account.

\section{Results}
Table~\ref{tab:probing_task_results_tacred} and Table~\ref{tab:probing_task_results_semeval} report the accuracy scores of the probing task experiments for models trained on the TACRED and SemEval dataset.  %
We did not include the \textbf{ArgOrd} and \textbf{EntExists} task in the SemEval evaluation, since SemEval relation arguments are always ordered in the sentence as indicated by the relation type, and entity types recognizable by standard tools such as Stanford CoreNLP that might occur between head and tail are not relevant to the dataset's entity types and relations.

\inserttacredprobingtable
\insertsemevalprobingtable
Baseline performances are reported in the top section of Table~\ref{tab:probing_task_results_tacred} and Table~\ref{tab:probing_task_results_semeval}. Length and ArgDist are both linear classifiers, which use sentence length and distance between \textit{head} and \textit{tail} argument as the only feature. BoE computes a representation of the input sentence by summing over the embeddings of all tokens it contains. Generally, there is a large gap between top baseline performance and that of a trained encoder. While SentLength and ArgDist are trivially solved by the respective linear classifier, BoE shows surprisingly good performance on SentLen and ArgOrd, and a clear improvement over the other baselines for named entity- and part-of-speech-related probing tasks.

\paragraph{Encoder Architecture}
For most probing tasks, except SentLen and ArgOrd, a proper encoder clearly outperforms bag-of-embeddings (BoE), which is coherent with the findings of \citet{adi_fine-grained_2017} and \citet{conneau_probing_2018}. Similarly, the results indicate that the prior imposed by the encoder architecture preconditions the information encoded in the learned embeddings. Models with a local or recency bias (CNN, BiLSTM) perform well on probing tasks with local focus, such as PosHead\{L,R\} and PosTail\{L,R\} and distance related tasks (ArgDist, ArgOrd).
Similarly, models with access to dependency information (GCN) perform well on tree related tasks (SDPTreeDepth). Due to the graph pruning step~\cite{zhang_graphco_2018}, the GCN is left with a limited view of the dependency tree, which explains the low performance on TreeDepth. Surprisingly, while Self-Attention exhibits superior performance on the RE task, it consistently performs lower on the probing tasks compared to the other encoding architectures. This could indicate Self-Attention encodes ``deeper'' linguistic information into the sentence representation, not covered by the current set of probing tasks.

\paragraph{Probing Tasks}
Compared to the baselines, all proper encoders exhibit consistently high performance on TypeHead and TypeTail, clearly highlighting the importance of entity type information to RE. In contrast, encoders trained on the downstream task perform worse on SentLen, which intuitively makes sense, since sentence length is mostly irrelevant for RE. This is consistent with~\citet{conneau_probing_2018}, who found SentLen performance to decrease for models trained on more complex downstream tasks, e.g.\ neural machine translation, strengthen the assumption that, as a model captures deeper linguistic properties it will tend to forget about this superficial feature. With the exception of the CNN, all encoders consistently show low performance on the argument distance (ArgDist) task. 
A similar performance pattern can be observed for ArgOrd, where models that are biased towards locality (CNN and BiLSTM) perform better, while models that are able to efficiently model long range dependencies, such as GCN and S-Att., show lower performance. The superior RE task performance of the latter indicates that their bias may allow them to learn ``deeper'' linguistic features. The balanced performance of CNN, BiLSTM and GCN encoders across part-of-speech related tasks (PosHeadL, PosHeadR, PosTailL, PosTailR) highlights the importance of part-of-speech-related features to RE, again with the exception of S-Att., which performs just slightly above baselines.
On TreeDepth and SDPTreeDepth (with GCN as the exception), average performance in many cases ranges just slightly above baseline performance, suggesting that TreeDepth requires more nuanced syntactic information, which the models fail to acquire. 
The good performance on grammatical role tasks (GRHead, GRTail) once more emphasizes the relevance of this feature to RE, with the GCN exhibiting the best performance on average. This is unsurprising, because the GCN focuses on token-level information along the dependency path connecting the arguments, and hence seems to be able to capture grammatical relations among tokens more readily than the other encoders (even though the GCN also does not have access to the dependency labels themselves).

\paragraph{Entity Masking}
Perhaps most interestingly, masking entity mentions with their respective named entity and grammatical role information considerably lowers the performance on entity type related tasks (TypeHead and TypeTail). This indicates that masking forces the encoder's focus away from the entity mentions, which is confirmed by the performance decrease in probing tasks with a focus on argument position and distance, e.g.\ ArgDist, ArgOrd, and SentLen. CNN and BiLSTM encoders exhibit the greatest decrease in performance, suggesting a severe overfitting to specific entity mentions when no masking is applied. In comparison, the GCN shows less tendency to overfit. Surprisingly, with entity masking the self-attentive encoder (S-Attn.) increases its focus on entity mentions and their surroundings as suggested by the performance increase on the distance and argument related probing tasks.

\paragraph{Word Representations}
Adding contextualized word representations computed by ELMo or BERT greatly increases performance on probing tasks with a focus on named entity and part-of-speech information. This indicates that contextualized word representations encode useful syntactic and semantic features relevant to RE, which is coherent with the findings of \citet{peters_deepcw_2018} and \citet{radford_improvinglu_2018}, who both highlight the effectiveness of linguistic features encoded in contextualized word representations (deep language representations) for downstream tasks. The improved performance on syntactic and semantic abilities is also reflected in an overall improvement in RE task performance. Compared to ELMo, encoders with BERT generally exhibit an overall better and more balanced performance on the probing tasks. This is also reflected in a superior RE performance, suggesting that a bidirectional language model encodes linguistic properties of the input more effectively.
Somewhat surprisingly, BERT without casing performs equally or better on the probing tasks focused on entity and part-of-speech information, compared to the cased version. While this intuitively makes sense for SemEval, as the dataset focuses on semantic relations between concepts, it is surprising for TACRED, which contains relations between proper entities, e.g. person and company names, with casing information more important to identify the entity type.

\paragraph{Probing vs. Relation Extraction}
One interesting observation is that encoders that perform better on probing tasks do not necessarily perform better on the downstream RE task. For example, CNN+ELMo scores highest for most of the probing tasks, but has an 8.1 lower F1 score than the best model on this dataset, S-Att.+BERT cased with masking. Similarly, all variants of the self-attentive encoder (S-Att.) show superior performance on RE but consistently come up last on the probing tasks, occasionally performing just above the baselines. ~\citet{conneau_probing_2018} observed a similar phenomena for encoders trained on neural machine translation.

\paragraph{Relation Extraction} 
The relation extraction task performance\footnote{See Appendix for more details on RE task performance, training, and model hyperparameters} on the TACRED dataset ranges between 55.3 (Bi-LSTM) and 57.6 F1 (S-Att.), with performance improving to around 58.8 - 64.7 F1 when adding pre-learned, contextualized word representations. As observed in previous work~\cite{zhang_position_aware_2017}, masking helps the encoders to generalize better, with gains of around 4 - 8 F1 when compared to the vanilla models. This is mainly due to better recall, which indicates that without masking, models may overfit, e.g.\ by memorizing specific entity names. The best-performing model achieves a score of 66.9 F1 (S-Att.+ BERT cased and masking). 

On the SemEval dataset performance of the vanilla models is around 80.0 F1. Adding contextualized word representations significantly improves the performance of all models, by 3.5 - 6 F1. The best-performing model on this dataset is a CNN with uncased BERT embeddings with an F1-score of 86.3, which is comparable to state-of-the-art models~\cite{wang_relation_2016,cai_bidirectional_2016}.

 \section{Related Work}
 \citet{shi_does_2016} introduced probing tasks to probe syntactic properties captured in encoders trained on neural machine translation. \citet{adi_fine-grained_2017} extended this concept of ``auxiliary prediction tasks'', proposing SentLen, word count and word order tasks to probe general sentence encoders, such as bag-of-vectors, auto-encoder and skip-thought. \citet{conneau_probing_2018} considered 10 probing tasks, including SentLen and TreeDepth, and an extended set of encoders such as Seq2Tree and encoders trained on NMT and NLI for general text classification. Their setup, however, is not directly applicable to relation extraction, because the RE task requires not only the input sentence, but also the entity arguments. We therefore extend their framework to accommodate the RE setting. Another difference to their work is that while their probing tasks focus on linguistic properties of general sentence encoders, we specifically focus on relation extraction. To that end, we extend the evaluation to relation extraction by introducing a set of 14 probing tasks, including SentLen and TreeDepth, specifically designed to probe linguistic properties relevant to relation extraction.

\section{Conclusion}
We introduced a set of probing tasks to study the linguistic features captured in sentence encoder representations trained on relation extraction. We conducted a comprehensive evaluation of common RE encoder architectures, and studied the effect of explicitly and implicitly provided semantic and syntactic knowledge, uncovering interesting properties about the architecture and input features. 
For example, we found self-attentive encoders to be well suited for the RE on sentences of different complexity, though they consistently perform lower on probing tasks; hinting that these architectures capture ``deeper'' linguistic features. We also showed that the bias induced by different architectures clearly affects the learned properties, as suggested by probing task performance, e.g.\ for distance and dependency related probing tasks. 

In future work, we want to extend the probing tasks to also cover specific linguistic patterns such as appositions, and also investigate a model's ability of generalizing to specific entity types, e.g.\ company and person names.

\section*{Acknowledgments}
We would like to thank all reviewers for their helpful comments and feedback. This work has been supported by the German Federal Ministry of Education and Research as part of the projects DEEPLEE (01IW17001) and BBDC2 (01IS18025E), and by the German Federal Ministry of Economics Affairs and Energy as part of the project PLASS (01MD19003E).

\bibliography{main}
\bibliographystyle{acl_natbib}

\appendix
\section{Further Training Details}

\subsection{Probing Task Training}
The probing task results reported in the main part are obtained by fitting a Logistic Regression classifier to the binary and multi-class classification task. We tune the $l_2$ penalty of the classifier with grid-search on the validation set.

\subsection{Relation Extraction Training}
For vanilla models we use 300-dimensional pre-trained GloVe embeddings~\cite{pennington_glove_2014} as input. Variants with ELMo use the contextualized word representations in combination with GloVe embeddings and models with BERT only use the computed representations. For models trained on TACRED we use 30-dimensional positional offset embeddings for head and tail (50-dimensional embeddings for SemEval). Similar for the batch-size we use 50 on TACRED and 30 on SemEval. If not mentioned otherwise, we use the same hyperparameters for models with and without entity masking.

\subsubsection{Hyperparameters}

\paragraph{CNN}
For training on TACRED we use the hyperparameters of \citet{zhang_position_aware_2017}. We employ \textit{Adagrad} as an optimizer, with an initial learning rate of 0.1 and run training for 50 epochs. Starting from the 15th epoch, we gradually decrease the learning rate by a factor of 0.9. For the CNN we use 500 filters of sizes [2, 3, 4, 5] and apply $l_2$ regularization with a coefficient of $10^{-3}$ to all filter weights. We use tanh as activation and apply dropout on the encoder output with a probability of 0.5. We use the same hyperparameters for variants with ELMo. For variants with BERT, we use an initial learning rate of 0.01 and decrease the learning rate by a factor of 0.9 every time the validation F1 score is plateauing. Also we use 200 filters of sizes [2, 3, 4, 5].

On SemEval, we use the hyperparameters of \citet{nguyen_recnn_2015}. We employ \textit{Adadelta} with initial learning rate of 1 and run it for 50 epochs. We apply $l_2$ regularization with a coefficient of $10^{-5}$ to all filter weights. We use embedding and encoder dropout of 0.5, word dropout of 0.04 and 150 filters of sizes [2, 3, 4, 5]. For variants using BERT, we decrease the learning rate by a factor of 0.9 every time the validation F1 score is plateauing.

\paragraph{BiLSTM}
For training on TACRED we use the hyperparameters of \citet{zhang_position_aware_2017}. We employ \textit{Adagrad} with an initial learning rate of 0.01, train for 30 epochs and gradually decrease the learning rate by a factor of 0.9, starting from the 15th epoch. We use word dropout of 0.04 and recurrent dropout of 0.5. The BiLSTM consists of two layers of hidden dimension 500 for each direction. For training with ELMo and BERT we decrease the learning rate by a factor of 0.9 every time the validation F1 score is plateauing.

On SemEval we instead use two BiLSTM layers with hidden dimension 300 for each direction, and also use embedding and encoder dropout of 0.5.

\paragraph{GCN}
On TACRED and SemEval we reuse the hyperparameters of \citet{zhang_graphco_2018}. We employ \textit{SGD} as optimizer with an initial learning rate of 0.3, which is reduced by a factor of 0.9 every time the validation F1 score plateaus. We use dropout of 0.5 between all but the last GCN layer, word dropout of 0.04, and embedding and encoder dropout of 0.5. Similar to the authors we use path-centric pruning with K=1. On TACRED we use two 200-dimensional GCN layers and similar two 200-dimensional feedforward layers with ReLU activation, whereas on SemEval we instead use a single 200-dimensional GCN layer.

\paragraph{Self-Attention}
After hyperparameter tuning we found 8 layers of multi-headed self-attention to perform best. Each layer uses 8 attention heads with attention dropout of 0.1, keys and values are projected to 256 dimensions before computing the similarity and aggregated in a feedforward layer with 512 dimensions. For training we use \textit{Adam} optimizer with an initial learning rate of $10^{-4}$, which is reduced by a factor of 0.9 every time the validation F1 score plateaus. In addition we use word dropout of 0.04, embedding dropout of 0.5, and encoder dropout of 0.5.

\section{Relation Extraction Results}
Table~\ref{tab:test_performance_semeval} and Table~\ref{tab:test_performance_tacred} show the relation extraction performances we obtained after training our model variants on the SemEval and TACRED dataset, respectively.

\insertsemevalresulttable

\inserttacredresulttable

\end{document}